# Is English the New Programming Language? How About Pseudo-code Engineering?

Gian Alexandre Michaelsen [a]
Renato P. dos Santos [a]

[a] Universidade Luterana do Brasil (Ulbra), Programa de Pós-Graduação em Ensino de Ciências e Matemática (PPGECIM), Canoas, RS, Brasil

## ABSTRACT

**Background:** The integration of artificial intelligence (AI) into daily life, particularly through chatbots utilizing natural language processing (NLP), presents both revolutionary potential and unique challenges. This research is motivated by the intricacies of human-computer interaction within the context of conversational AI, focusing on the role of structured inputs from pseudo-code engineering in enhancing chatbot comprehension and response accuracy. **Objectives:** This study investigates the comparative effectiveness of natural language versus pseudo-code engineering generated inputs in eliciting precise and actionable responses from ChatGPT, a leading language model by OpenAI. It aims to delineate how different input forms impact the model's performance in understanding and executing complex, multi-intention tasks. **Design:** Employing a case study methodology supplemented by discourse analysis, the research analyzes ChatGPT's responses to inputs varying from natural language to pseudo-code engineering. The study specifically examines the model's proficiency across four categories: understanding of intentions, interpretability, completeness, and creativity. **Setting and Participants:** As a theoretical exploration of AI interaction, this study focuses on the analysis of structured and unstructured inputs processed by ChatGPT, without direct human participants. **Data collection and analysis:** The research utilizes synthetic case scenarios, including the organization of a "weekly meal plan" and a "shopping list," to assess ChatGPT's response to prompts in both natural language and pseudo-code engineering. The analysis is grounded in the identification of patterns, contradictions, and unique response elements across different input formats. **Results:** Findings reveal that pseudo-code engineering inputs significantly enhance the clarity and determinism of ChatGPT's responses, reducing ambiguity inherent in natural language. Enhanced natural language, structured through prompt engineering techniques, similarly improves the model's interpretability and creativity. **Conclusions:** The study underscores the potential of pseudo-code engineering in refining human-AI interaction, advocating for its broader application across disciplines

Corresponding author: Renato P. dos Santos. Email: renatopsantos@ulbra.edu.br

requiring precise AI responses. It highlights pseudo-code engineering's efficacy in achieving more deterministic, concise, and direct outcomes from AI systems like ChatGPT, pointing towards future innovations in conversational AI technology.

## INTRODUCTION

The integration of artificial intelligence (AI) into our daily lives has not only revolutionized our interaction with technology but also posed unique challenges, particularly in the realm of natural language processing (NLP). As AI systems, like chatbots, become more sophisticated, they increasingly rely on NLP to interpret and respond to human language with a level of nuance previously unattainable. As an example, we can see a input text for an AI, also known as a *prompt,* developed from Mollick & Mollick in natural language that aims to use the AI to generate examples for the explanation of complicated concepts.

> "I would like you to act as an example generator for students. When confronted with new and complex concepts, adding many and varied examples helps students better understand those concepts. I would like you to ask what concept I would like examples of, and what level of students I am teaching. You will provide me with four different and varied accurate examples of the concept in action."(2023, 5-6).

Independently the clarity of the written prompt, the inherent ambiguity and complexity of natural language can often hinder these systems' ability to decipher user intent accurately (Carvalho, 2021; Jho, 2020). In the utilization of elaborated prompts with multiple functions, we notice that, in some cases and moments, the ChatGPT had opened window where the requested process was processed thru a programming language. By noticing some nuances between the requisition and what ChatGPT was doing in the execution we delve ourselves in approximate the natural language to programming languages that could be executable to ChatGPT. This is where the concept of pseudo-code emerges as a potential bridge between the intuitive, flexible nature of human language and the precise, logical structure of programming languages. Pseudo-code, an informal high-level description of programming logic, offers a solution to the challenges posed by natural language's ambiguity. It provides a structured yet readable format that can significantly enhance the clarity of instructions for Large Language Models (LLMs) like ChatGPT. By incorporating elements of pseudo-code into user prompts, we can minimize misunderstandings and improve the effectiveness of AI interactions. This approach aligns closely with the latest advancements in Generative AI, which seeks not only to interpret data but to create original content, underscoring the need for clear communication between humans and AI systems (Jovanovic & Campbell, 2022).

In the spectrum of languages used with advanced models like ChatGPT, pseudo-code can act as a new communication tool, bridging natural language and formal programming. Pseudo-code is an informal, high-level description method for problem-solving solutions, aiding programmers in algorithm conception before implementation in languages like C#, C++, or Java. This approach enhances logical reasoning and is accessible to non-experts, making it invaluable in language model interactions (Oda et al., 2015).

Here bellow we can see a briefly demonstration of the pseudo-code format, as an example we can take a ChatGPT prompt showed before and rebuild it as a pseudo-code, with give him this following structure.

> 1) ACT as an example generator for students to help them to better understand those concepts.
>
> 2) ASK what concept I would like examples of.
>
> 3) ASK what level of students I am teaching.
>
> 4) IF confronted with new and complex concepts, THEN add many and varied examples
>
> 5) GENERATE four different and varied accurate examples of the concept in action.

As we can see, the same requisition made the prompt developed by (Mollick & Mollick, 2023) can be pseudo-coded, as like any other requests that the user can make to a chatbot. Based on this, our study utilizes pseudo-code to explore the intersection of computational logic and human language flexibility, examining how ChatGPT processes and responds to semi-algorithmic commands. Incorporating chain of thought techniques (Ling et al., 2023; Wei et al., 2022) and prompt engineering (Heston & Khun, 2023), and the some of the logics of codes scription, pseudo-code could became a powerful tool for effective communication with language models, offering insights into human-computer interaction.

Our study delves into the comparative effectiveness of traditional natural language prompts versus those structured as pseudo-code and enhanced pseudo-code in eliciting accurate responses from ChatGPT. Employing a case study methodology (Yin, 2009) coupled with discourse analysis (Jørgensen & Phillips, 2011), we aim to illuminate the nuances in how ChatGPT processes these different forms of input. This investigation is motivated by the hypothesis that pseudo-code, by reducing the ambiguity inherent in natural language, can lead to more precise and actionable responses from LLMs.

The article is methodically structured to guide the reader through our exploration of this hypothesis. Beginning with an overview of the operational principles of NLP, we transition to a focused examination of pseudo-code—its definition, significance, and application in improving LLM interactions (Calzadilla, 2018; Oda et al., 2015). Subsequent sections detail the methodological framework adopted for this study, including the creation of pseudo-codes and the analytical approach to evaluating ChatGPT's responses. By juxtaposing the performance of traditional natural language prompts with those framed as pseudo-code, we shed light on the potential of structured inputs to refine and elevate our engagement with AI technologies.

## THEORETICAL FRAMEWORK

### Natural Language Understanding

Natural Language Understanding (NLU), a cornerstone of Natural Language Processing (NLP), activates when a user engages with a chatbot. This initial phase is pivotal, enabling machines to analyze and decipher language by identifying concepts, emotions, entities, and keywords, all while zeroing in on the user's intention. NLU's role is paramount in customer service applications, where grasping and addressing verbally or textually reported issues is necessary (Khurana et al., 2022).

*Exploring the Levels of Human Language in NLU*

**Phonology:** Understanding the sound organization in languages is essential. For instance, the phonological nuances in the pronunciation of "read" can alter its meaning based on the tense, thereby affecting the message's interpretation in an audio interpretation on a chatbot interaction. A practical application is seen in virtual assistants like Siri or Alexa, which analyze vocal inputs to discern contextually relevant meanings of homonyms based on the user's query.

**Morphology:** Examines word structures and morphemes, the smallest meaning units. Analyzing morphological variations helps interpret different contextual uses, like understanding "running" as an action or a continuous state in user queries. For example, chatbots in fitness apps utilize this understanding to differentiate between "running as exercise" versus "running a program" based on the user's activity and preferences.

**Lexical:** Delves into the meanings of words, crucial for resolving ambiguities. For example, a chatbot discerning the word "bat" as sports equipment or an animal based on the conversation context. E-commerce chatbots exemplify this by accurately suggesting products (a baseball bat) or information (about bats) depending on the surrounding text in user inquiries.

**Syntactic:** Entails grammatical sentence analysis, key to understanding word order and relationships, such as distinguishing between "eating apples" and "apples eating" scenarios. Customer support chatbots use syntactic analysis to correctly interpret requests, ensuring responses are accurate and relevant to the query's structure.

**Semantic:** Aims to comprehend the actual meanings of texts or sentences, processing the logical structure to grasp the interactions among words or concepts, like differentiating "I saw her duck" based on action or observation. Translation services in messaging platforms showcase this level by providing contextually appropriate translations that consider both literal and figurative language use.

**Pragmatic:** Focuses on implied meanings, considering knowledge beyond the text. This layer allows NLU to infer that a user asking, "Can you open the window?" is requesting an action, not querying the chatbot's capabilities. Smart home assistants demonstrate pragmatics by interpreting commands in the context of the home environment, enabling them to perform tasks like adjusting a thermostat based on conversational cues rather than explicit instructions.

*Challenges and Solutions in NLU*

One of the principal challenges in Natural Language Understanding (NLU) is the presence of linguistic ambiguities that can lead to varied interpretations of a sentence or phrase. These ambiguities are addressed at different linguistic levels:

> **Syntactic Ambiguity:** A common issue where the structure of a sentence allows for multiple interpretations. Consider the phrase "Visiting relatives can be annoying"; it can be understood in two ways: either the act of visiting relatives is annoying, or the relatives who visit are annoying. NLU syntactic analysis tools strive to clarify such ambiguities. For instance, an NLU system might use user interaction history or follow-up questions to discern if the user generally discusses family matters negatively or positively, guiding the system towards the intended interpretation.
>
> **Semantic Ambiguity:** Occurs when a word or phrase has multiple meanings, leading to confusion about the intended message. The word "crane" can refer to a bird, a type of machinery for lifting, or an action to stretch the neck. Semantic clarification techniques in NLU determine the correct meaning by analyzing the surrounding text or context. For example, if a user is discussing construction, the NLU system is more likely to interpret "crane" as machinery, showcasing the strategy of preservation by keeping the contextually relevant meaning.
>
> **Lexical Ambiguity:** Deals with words that have multiple meanings, leading to confusion without context. The sentence "I went to the bank" is ambiguous without additional information. Lexical disambiguation methods in NLU use context clues from the conversation, like recent transactions or geographical location, to deduce whether "bank" refers to a financial institution or riverbank. This is an application of minimization, where the system reduces ambiguity by leveraging known context.

In addressing these challenges, NLU systems employ strategies like **minimization** (to reduce the chances of ambiguity), **preservation** (to maintain the intended meaning as closely as possible), and **interactive disambiguation** (engaging in a dialogue with the user to clarify ambiguous statements in real-time), as outlined by Khurana et al. (2022). An example of interactive disambiguation is when a chatbot, unsure of a user's intent due to ambiguous input, asks clarifying questions to ensure accurate understanding and response, thus enhancing the chatbot's ability to comprehend and engage effectively.

Furthermore, incorporating cognitive comprehension tasks such as sentiment analysis and emotion detection allows conversational agents to better grasp the tone and underlying feelings in user communications. This might manifest in a customer service chatbot that, detecting frustration in a user's message about "waiting at the bank," understands the sentiment as negative and prioritizes a response designed to alleviate frustration, enriching the interaction experience (Kusal et al., 2022).

*Action Orientation in NLU*

To enrich the understanding of Action Orientation in Natural Language Understanding (NLU), it's crucial to delve deeper into the mechanism of converting textual data into N-dimensional feature vectors or word embeddings (Wolfram, 2023). This transformative process is at the heart of NLU, enabling the detailed semantic representation of language inputs. Consider the task of processing a user's query, such as "Find me a quiet café nearby." NLU systems tackle this by breaking down the query into vectors that encapsulate the essence of "quiet," "café," and "nearby," thereby allowing the model to understand and categorize the user's intent with high precision.

These vectors serve as the foundation for intention classification models within NLU frameworks, crucial for the operation of sophisticated technologies like Large Language Models (LLMs)(Wolfram, 2023). Through vectorization, NLU facilitates an enhanced understanding of language nuances, significantly improving the accuracy and relevance of responses generated by chatbots in real-time interactions. This intricate process underscores the advanced capabilities of NLU in interpreting and acting upon human language, providing a seamless interface for human-chatbot communication (Kusal et al., 2022).

This detailed exploration of action orientation in NLU showcases the essential role of word embeddings in enhancing the interpretative capabilities of AI systems, ensuring that user queries are met with precise and contextually relevant responses.

**Pseudo-code in the Context of ChatGPT and Large Language Models Interactions**

Pseudo-code emerges as a pivotal tool in bridging the inherent challenges of Natural Language Understanding (NLU) with the structured logic required for programming, as highlighted by Calzadilla (2018). It serves as a clarifying medium between the ambiguous flexibility of human language and the deterministic precision of programming languages, exemplifying the interdisciplinary approach central to this research. By converting a natural language prompt by Mollick & Mollick (2023) into pseudo-code, this methodology illuminates how structured action sequences can directly address NLU's obstacles, enhancing ChatGPT's response efficacy. The use of explicit programming constructs such as conditional statements ("IF", "THEN") and commands ("ASK", "GENERATE"), emphasized in uppercase, signals precise instructions to the LLM, thereby reducing the ambiguity that NLU systems often grapple with.

This application of pseudo-code not only refines the functionality of LLMs but also embodies the interdisciplinary nexus of linguistics, computer science, and AI research. It showcases how integrating simple programming logic into the instruction of AI models can mitigate some of the most persistent challenges in NLU, including the disambiguation of user intent and the contextual interpretation of queries. Consequently, pseudo-code becomes not just a tool for enhancing AI interactions but also a demonstration of how the confluence of diverse scientific disciplines can lead to innovative solutions in AI technology (Calzadilla, 2018). This research underscores the value of pseudo-code in transcending traditional boundaries between natural language and programming, offering a comprehensive approach to tackling the complexities of human-computer interaction.

### Large Language Model (LLM)/ChatGPT

*From NLU to Vectorization*

The journey of processing language in ChatGPT, a distinguished Large Language Model (LLM), begins with an essential phase known as Natural Language Understanding (NLU). In this initial stage, the model undertakes the task of interpreting the intricacies of human language. The process progresses into a critical phase termed vectorization or 'embedding', where the essence of language transitions from a textual to a numerical form. This transformation is pivotal, concentrating on encapsulating elements that are central to the user's intention. ChatGPT accomplishes this through the utilization of 'tokens', which are numerical representations of words or text fragments. These tokens are meticulously selected based on their proficiency in capturing the core content and the underlying intent of the input. The selection process is comprehensive, incorporating an evaluation of the contextual relevance and semantic importance of each word or phrase, ensuring that the most significant aspects of the language are retained (Wolfram, 2023).

In the subsequent phase, these tokens undergo a mapping process to become vectors, creating 'embeddings' that are imbued with rich semantic and contextual properties. These embeddings are then integrated into the neural network, essentially influencing the model's response mechanisms to user commands. It's important to note that while the structural integrity of the network remains intact during real-time processing, this mechanism enables ChatGPT to offer dynamic interpretations and responses to a wide array of human language inputs. The selection and mapping of tokens to vectors are not arbitrary; instead, they are governed by sophisticated algorithms designed to understand the nuanced relationships between words in different contexts. This ensures that the embeddings reflect a deep understanding of language nuances, enabling the neural network to process requests with a high degree of accuracy and relevance (Wolfram, 2023). Once selected, these tokens are mapped to vectors, forming 'embeddings' rich in semantic and contextual attributes, and fed into the neural network. These embeddings dictate the neural network's response to user commands. While the network's structure remains unchanged in real-time processing, this approach allows ChatGPT to interpret and respond dynamically to diverse human language inputs (Wolfram, 2023).

*From Vectorization to Neural Network Training*

Large Language Models (LLMs) are advanced systems that process enormous volumes of data to solve complex tasks and problems. They go beyond being mere repositories of information; they are sophisticated structures for understanding and generating language. Imagine a conceptual space where information is organized in distinct layers, like in a vast archive (Wolfram, 2023).

When input data (in the form of embeddings) arrive at an LLM, they go through several layers of processing in the neural network. Each layer of the network specializes in different aspects of language, such as semantics, grammar, or context. The role of vectors here is crucial: they guide the model to relevant parts of the 'conceptual space' of stored knowledge. This space is an abstract representation of all the data that the model has been trained to understand (Wolfram, 2023).

For example, when creating ice cream recipes, the vectors direct the model to parts of the conceptual space related to ice creams. If the request specifies 'lactose-free ice cream recipes', the system adjusts its vectors to focus on the intersection between ice creams and lactose-free alternatives. This dynamic process of vector adjustment is what allows the model to respond flexibly and relevantly to user queries.

In this conceptual space, the vectors act as guides, highlighting relevant information and temporarily relegating less pertinent data to the background. This dynamic serve to illustrate how a neural network reorganizes and prioritizes data in response to a specific query. Each user interaction prompts the neural network to traverse this conceptual space, adapting to highlight relevant information. This adaptive process is akin to 'network training' or 'machine learning', where the model refines its responses based on user input, analogous to human learning (Wolfram, 2023).

Similarly to human training, the machine may not initially act as desired by the user. Internally, the layers of the network adjust through vectors to achieve a closer result to the desired one. During this training, the model generates different examples to refine its response, learning and shaping itself according to the user's needs (Wolfram, 2023).

*Neural Networks: Managing Information*

ChatGPT's neural networks are characterized by attributes such as non-linearity, unlimited capacity, adaptability, self-organization, and non-convexity. These attributes are not just technical jargon; they are the backbone of the model's ability to handle complex tasks, adapt to varied inputs, and offer a variety of solutions and responses. They are what make the network adept at learning, storing associative memories, and finding optimal solutions. The process allows the model to dynamically adjust its outputs based on the nuances of each query, ensuring that responses are not only accurate but also contextually aligned with the user's expectations. This adaptability is a testament to ChatGPT's advanced understanding of language and its capacity to provide tailored responses across a spectrum of inquiries. This occurs thanks to the highly strategic selection process, which consider the semantic weight and contextual relevance of each word to be converted into tokens. This versatility is crucial, not just for tasks like image recognition and predictions, but also for interpreting and responding to natural language and pseudo-code input, each with its unique challenges and requirements (Kusal et al., 2022; Wolfram, 2023; Wu & Feng, 2018).

When faced with complex queries, for example, in responding to a multifaceted question about climate change's impact on global agriculture, ChatGPT would dissect the query into components such as 'climate change effects' and 'global agriculture'. It then applies its vectorized knowledge to each segment, drawing on its vast training data to assemble a comprehensive response that addresses each aspect of the query. This process not only ensures accuracy but also enriches the response with a breadth of insight and depth that might not be immediately apparent in the query itself.

The model's handling of complex queries and its adaptive response mechanism underscore its sophistication. By leveraging detailed embeddings and an intricate understanding of context, the LLM navigates the complexities of human language with agility. This nuanced approach enables the model to offer responses that are not just reactive but deeply informed, providing users with insights that are both precise and contextually rich (Wolfram, 2023).

The evolution to deeper neural networks, a foundational aspect of Deep Learning, has significantly amplified the model's ability to process vast and intricate data sets. This enhanced processing capability is directly linked to the model's performance in interpreting both natural language and pseudo-code. While natural language demands an understanding of nuanced human expression, pseudo-code requires a more structured and logical interpretation. The neural network's multi-layered architecture equips ChatGPT to navigate these diverse linguistic landscapes, extracting and responding to information in a way that is representative, valuable, and, most importantly, aligned with the user's intent (Wu & Feng, 2018).

*The theorical difference to a LLM deal with a Natural Language and a Pseudo-code prompt*

The exploration of LLMs like ChatGPT highlights a fundamental difference in how these models process natural language prompts compared to pseudo-code, each imposing distinct demands. The NLP relies on the model's capability to discern the intricacies of human communication, necessitating a flexible approach for accurately interpreting user intent (Wolfram, 2023). Conversely, pseudo-code, with its structural similarity to programming languages, requires ChatGPT to engage in a more analytical manner, demanding precise adherence to structured inputs to operationalize embedded logical instructions accurately. The utilization of pseudo-code can significantly enhance ChatGPT's output determinism and response clarity, allowing for outputs that can more accurately reflect the user's procedural logic and intention, thereby minimizing ambiguity.

The process of vectorization, crucial in transforming textual inputs into numerical vectors or embeddings, is instrumental across both domains but shines particularly in the domain of pseudo-code. This transformation can facilitate the mapping of structured commands to specific actions or responses, streamlining the interpretative process and enhancing the model's adaptability (Kusal et al., 2022; Wu & Feng, 2018). The neural network architecture underlying ChatGPT further augments this adaptability, enabling the model to adeptly navigate the interpretative demands of natural language and the logical rigors of pseudo-code. The application of pseudo-code can showcase the model's capacity for high-level problem-solving and logical reasoning, attributes that are significantly bolstered by the structured nature of pseudo-code inputs.

In the next section, we will explain the methodological framework chosen to conduct our case study and further demonstrate the evidence of the advantages that resulted from the use of pseudo-codes. Doing a comparative analysis, we provide empirical evidence supporting pseudo-code efficiency and clarity over natural language instructions. Expanding on the technical mechanisms utilized to write a pseudo-code, nominating this technique as Pseudo-code Engineer.

## METHODOLOGY

The present article incorporates a case study methodology (Yin, 2001), with the specific objective of investigating the peculiarities and unique challenges presented by ChatGPT in the context of the interaction between natural language and pseudo-code. ChatGPT, being an advanced language model developed by OpenAI, offers a unique case study scenario due to its language processing capabilities and flexibility in dealing with different input formats. This case study will focus on uncovering how ChatGPT interprets and responds to commands in natural language compared to pseudo-code. The choice of ChatGPT as a case study is justified by its emerging relevance in various practical applications and the need to deeply understand its capabilities and limitations in a complex interactive scenario.

This case study methodology, as proposed by Yin (2001), allows for a deep and contextual analysis of the case, which is the interaction with ChatGPT. The choice of this approach is justified by the complexity and multifaceted nature of the phenomenon under study, the interaction between natural language and programming language, and ChatGPT's ability to process and respond to these two distinct forms of communication. The study will specifically focus on the nuances of this interaction, exploring how ChatGPT interprets and translates commands in natural language compared to pseudo-code. This will include a detailed analysis of the processes of intention interpretation, and the responses generated by the model in different scenarios and types of inputs.

The interpretation of the data will be carried out through the methodology of discourse analysis (Jørgensen & Phillips, 2011). This methodology offers a deep and reflective approach, allowing for a holistic understanding of the phenomenon studied. Discursive textual analysis is not limited to the mere decomposition of texts; it involves an iterative and critical process of reading, interpretation, and reinterpretation of the data.

Specifically, this methodology will be employed to explore and understand the interactions between the user and ChatGPT. It will enable the identification of patterns, contradictions, and nuances in the responses generated by the system in different input formats, whether they be natural language or pseudo-code. The process of discursive textual analysis in this research will include data collection in the form of textual interactions with ChatGPT, extracted into a text file for subsequent coding and categorization of these data. Later, the categorized data will be subjected to a critical analysis, where the results brought by the LLM will be examined. This approach will allow for a deeper understanding of the internal mechanisms and characteristics of ChatGPT, as well as the practical implications of its use in different contexts.

The methodology adopted in this study, which encompasses both the case study and the discourse analysis, has been carefully chosen to align directly with the specific research objectives. The case study, focused on ChatGPT, provides a lens through which we can explore and deeply understand the nuances of the interaction between natural language and pseudo-code. This method allows us to investigate in a detailed and contextualized manner how ChatGPT processes and responds to different input formats, in line with our main research questions (Yin, 2001).

> *How does ChatGPT handle multiple intentions inputted in the form of natural language?* We will analyze the interactions to understand how ChatGPT identifies and responds to various intentions naturally expressed by the user.

*How does ChatGPT deal with multiple intentions when inputted in the form of pseudo-code?* We will examine whether the clarity and structure of the pseudo-code influence the accuracy and determinism of ChatGPT's responses compared to natural language inputs.

We assume as a proposition the conception that:

- The use of prompts with multiple intentions in natural language can lead to results that diverge in an indeterministic way from the user's intended requests.
- The use of pseudo-code may facilitate the identification of intentions recognized by the LLM.
- The possibility of increasing the determinism in the interactions that follow the use of pseudo-code.

The analyses will be structured around three distinct units, each focusing on a different method of interaction with ChatGPT. These units are:

**Natural Language (Unit A)**: This unit will consist of interactions carried out using natural language. The focus will be on how ChatGPT interprets and responds to inputs formulated in common human language, identifying patterns and nuances in the model's responses.

**Enhanced Natural Language (Unit B):** This unit delves into the realm of interactions facilitated through an advanced variant of natural language, meticulously crafted to augment the precision and lucidity of user commands directed at ChatGPT. Unlike conventional natural language inputs, this enhanced form incorporates sophisticated prompting techniques designed to refine and specify user intents, thereby facilitating a more accurate and nuanced response from the language model. The enhancement of natural language encompasses strategic modifications such as the inclusion of specific contextual cues, explicit instruction framing, and the integration of prompts that mimic a chain of thought process, aiming to guide the model through a logical sequence of reasoning that mirrors human cognitive processes. Examples of such enhancements include structuring prompts to include step-by-step questions or scenarios that lead ChatGPT to deduce the user's intent more effectively, as well as embedding emotional or situational context to elicit responses that are not only relevant but also empathetic and context-aware. These techniques, derived from recent advancements in natural language processing and AI interaction studies (Heston & Khun, 2023; Li et al., 2023; Ling et al., 2023a, 2023b; Wei et al., 2022), empower users to elicit more detailed, accurate, and contextually rich responses from ChatGPT, illustrating the potential of enhanced prompts in bridging the gap between human queries and AI comprehension. The subsection "Enhancement of Prompt in NL" provides a comprehensive overview of these processes, detailing the methods employed to elevate the standard of natural language communication with ChatGPT, thereby setting a precedent for improved interaction quality and model responsiveness.

**Pseudo-code (Unit C):** This unit will concentrate on interactions carried out through pseudo-codes, allowing the evaluation of how ChatGPT responds to a more structured and directed form of input, and whether this results in more accurate or relevant responses. The methodology for enhancing pseudo-code entails the integration of specific programming concepts, such as conditional statements, loops, and function calls, designed to test the language model's ability to follow complex instructions and execute logical sequences.

Incorporating elements such as detailed comments to guide the model's reasoning, explicitly defining variables and their expected outcomes, and structuring the pseudo-code to reflect the sequential logic found in software development practices, aims to provide ChatGPT with a clear, unambiguous set of instructions that require a higher level of cognitive processing. This approach not only tests the model's capacity to parse and execute more complex instructions but also its ability to engage with the underlying logic of programming tasks, offering insights into its computational thinking capabilities.

We will further present the section which detail the actions they can be taken to improve pseudo-codes. Thus techniques drawing from contemporary research in programming language theory and artificial intelligence interaction (Heston & Khun, 2023; Wei et al., 2022)(Yue et al., 2023), to offer a granular view of how structured, logical enhancements can significantly impact ChatGPT's interpretative and problem-solving performance.

To conduct an effective comparison between the units of analysis, we will adopt a systematic approach:

**Parallelism in Interactions:** We will ensure that the interactions in both units are parallel in terms of intentionality and context. This means that the same types of requests or commands will be presented to ChatGPT, first in natural language and then in pseudo-code.

**Comparative Analysis:** We will use discourse analysis to examine ChatGPT's responses in the units. This will include identifying differences and similarities in the responses, the accuracy of the model's interpretations, and the relevance of the given responses.

**Interpretation of Results:** We will interpret the results to understand the implications of the form of input on the effectiveness of ChatGPT's response. This will help us determine whether the use of pseudo-codes offers a significant advantage in terms of clarity and accuracy in interaction with the model.

**Data collection process**

Thru the Chat GPT 4 we inserted the respective entry, in each format that we describe above in the units A, B, C and D. In the "Unit A" we only we had one single interaction since the prompt was already fulfilled with the parameters to perform solicited task. In the following units, as a demanding characteristic of the generalization process, we had to had two interactions with the LLM. The first one only caring the text or the pseudo-code to describe and orientate the LLM to perform the desired task and a second one to fulfill the LLM with the already selected parameters. To only analyze the coding differences described in each unit, we standardize the parameters interaction, utilizing the sabe format in the units B and C. It's possible to see the parameters interaction patterns in the nexus A, B and C.

Through this comparative method, grounded in discourse analysis (Jørgensen & Phillips, 2011), we establish clear objectives to evaluate ChatGPT's responses in-depth. The specific objectives include:

- Exploring the accuracy of ChatGPT in understanding user intentions in both natural language and pseudo-code inputs, identifying how the model interprets and responds to these two distinct formats.

- Investigating the consistency of ChatGPT's responses concerning the complexity and ambiguity of inputs, identifying how the model deals with requests that have multiple intentions or are structurally complex.

- Examining the differences in ChatGPT's responses when confronted with inputs in natural language compared to pseudo-

codes, seeking to understand how the structure and form of the input influence the response generation.

Discourse analysis (Jørgensen & Phillips, 2011) complements the approach of this case study (Yin, 2001), allowing us to uncover deeper layers of meaning in ChatGPT's responses. This conjunction of methodologies is essential for the critical and detailed interpretation of the collected data, enabling us to identify patterns and trends in the model's responses. Thus, the synergy between the case study and discourse analysis equips us to approach our research questions with a robust and integrated methodology, ensuring that each aspect of the interaction with ChatGPT is explored in a comprehensive and reflective manner.

**Analysis steps**

The textual discourse analysis conducted in this study was meticulously designed to dissect the nuanced interactions between users and ChatGPT across various linguistic formats. This analytical process, deeply rooted in the principles of discourse analysis as outlined by (Jørgensen & Phillips, 2011), was bifurcated into distinct phases: codification and categorization, each aimed at unraveling the complexities of language model interactions.

*Codification Process*

The codification process commenced with a thorough examination of the responses generated by ChatGPT under different linguistic inputs: natural language, enhanced natural language and pseudo-code. Each response was meticulously dissected to identify and extract key themes, patterns, and linguistic nuances, aligning with the study's objectives to compare the model's proficiency across varied inputs. This phase was instrumental in breaking down the textual data into manageable units for deeper analysis, allowing for the isolation of significant elements that denote the model's interpretative capabilities and response accuracy.

*Categorization Framework*

After codification, the categorization phase involved the systematic grouping of coded data into distinct categories, each representing a core aspect of ChatGPT's interaction with the input languages. Drawing from the foundational methodologies of Jørgensen and Phillips (2011), four primary categories were established:

1. **Understanding of Intentions:** Assess the ability of ChatGPT to identify and respond to multiple intentions contained in a single entry. This category will focus on analyzing whether the model can discern and address all aspects requested in a complex prompt.

2. **Interpretability**: The focus here is on understanding whether ChatGPT's responses are presented in a way that users can comprehend the reasoning behind the provided solutions, particularly in contexts that require step-by-step problem resolutions. The emphasis is on the clarity of communication and the ease of understanding of the responses.

3. **Completeness:** This category assesses the extent to which ChatGPT can cover all the essential aspects of a meal plan. The focus is on the model's ability to include each relevant element, ensuring that the response is comprehensive and meets all the user's needs and requirements.

4. **Creativity:** This dimension assesses the degree of organizational creativity (Lee & Choi, 2003) manifested in the meal suggestions proposed by each analytical unit, with a particular focus on the diversity and originality of the options generated by the AI system. Through this evaluation, it becomes possible to ascertain the effectiveness with which ChatGPT integrates and reconfigures nutritional and culinary knowledge to forge innovative ideas and solutions.

These categories were not merely descriptive but analytical, facilitating a nuanced understanding of the language model's operational dynamics. Each category has their specific criteria where the units will be evaluated. Such evaluation will indicate if the present unit "Fully meet", "Partially meet" or "Not meet" the respective criteria. The criterions are essential to evaluate how each unit performs in each category and could be helpful to identify their possible fragilities.

Regarding the qualitative analyses of the collected materials, we developed specific criteria to each one of the categories. Each unit of analysis was analyzed though each criterion to verify if it attempts, partially attempt, or not attempt at these specific criterions. To organize the criterions in their categories we code them with a number, meeting the category he belongs and a letter to distinguish either one. Down here we present the code, the name of the criterion and respectively their explanation.

> **1.A: Understanding User Goals**: This includes the model's ability to recognize the main goal of the request, such as creating a Paleo meal plan to gain lean mass within a specific budget.
>
> **1.B: Budget Adaptation:** Considers the model's ability to include the budget factor in its responses, whether through listing prices or providing tips for managing costs efficiently.
>
> **2.A: Structure and Coherence of Responses:** Evaluates exclusively on the organizational and logical aspects of response presentation. Highlight the necessity for coherence, clear transitions, and logical progression to ensure responses are intuitively structured and easy to follow.
>
> **2.B: Information to Support Informed Decisions:** Considers whether the information given effectively supports the user in making informed and conscious choices. This entails an assessment of how well nutritional information, and recommendations are integrated in a way that educates the user about the value of each choice.
>
> **3.A: Comprehensive Coverage of Needs:** The analysis focuses on how well each unit addresses all relevant needs within the category in question. In this case, it is checked that the meal plans cover all main meals and snacks for each day of the week, ensuring that there are no important omissions that could affect the user's daily diet.
>
> **3.B: Daily Meal and Snack Inclusion:** Checks for the inclusion of all meal types (breakfast, lunch, dinner, snacks) for each day, ensuring a comprehensive daily diet plan.
>
> **4.A Meals Development Creativity:** Observes the diversity and quantity of proposed meal options, focusing on lean proteins, healthy fats, and a selection of vegetables and fruits, in alignment with the user's health and fitness goals. In these criteria we assume that the unites that creates the biggest variety of meals will fully meets the criteria, the units that developed in the middle range will partially meets and the unit that build up less meals, in comparison with the others will be evaluate with the designation of not meeting the criteria.

**4.B Adaptability and Flexibility of Meal Plan:** Evaluates suggestions for substituting ingredients or meals to maintain variety and interest, demonstrating flexibility in meal planning. Assesses the model's creativity in offering adaptable meal plans that can be customized according to individual preferences, seasonal availability of ingredients, or last-minute budget adjustments.

To the criterions that involved counting, like the criterion 4.A, the units with the highest counts were denominated as "fully meeting the criteria" and respectively, the units with the lower counts got the denomination of "not attempting".

*Numerical Analytic process*

The assessment derived from each criterion will be quantified, assigning a numerical value to facilitate comprehensive conclusions for each category and within our overarching results. A value ranges from "1" to "3" will be employed, corresponding to the unit's fulfillment status: "1" signifies the unit's failure to meet the criterion, "2" indicates partial achievement, and "3" denotes full attainment. This methodical quantification will significantly aid in appraising the collective performance of each unit across all established categories.

*Compositive analysis*

The literature on discourse Analysis (Jørgensen & Phillips, 2011) steers the analyst towards a composite examination of outcomes to delineate distinctions among the materials scrutinized. Initially, the findings of each category are deliberated upon, leveraging the designated criteria for each. Subsequently, a holistic analysis is conducted, entailing an integrative comparison of each unit's cumulative performance across the entire evaluation, encompassing both criteria and categories.

**Prompts Development**

For the development of this research, which aims to study how ChatGPT handles complex interactions with multiple intentions, we will first define what we understand by this type of interaction. We identify complex interactions with multiple intentions as those that, through a single prompt, encompass different aspects or functions. It is expected that these actions will be executed in a single request through this prompt. In this way, we develop a synthetic case that can better exemplify what we desire.

An organization of a "weekly meal plan" and a "shopping list" were the selected tasks to execute our analysis of ChatGPT interactions with multiple intentions, we consider the complexity and multifaceted nature of meal planning as integral to understanding the depth and flexibility of ChatGPT. This task stands out as a particularly poignant test case due to its inherent requirement for understanding a diverse range of inputs, including dietary preferences, nutritional goal and the budget constraints. Such a scenario demands the model not only to grasp the explicit requests made by the user but also to infer underlying intentions and preferences that might not be directly stated.

The "weekly meal plan and shopping list" task encapsulates multiple layers of intention within a single inquiry, challenging ChatGPT to demonstrate its capacity for nuanced understanding, contextual interpretation, and creative problem-solving. This task requires the model to navigate through complex dietary requirements, optimize for health and budget considerations, and generate a coherent plan that aligns with the user's goals. By employing this task as a lens through which to examine ChatGPT's interactions, we aim to capture a broad spectrum of the model's capabilities, from basic comprehension to advanced reasoning and innovation. This approach allows us to meticulously dissect the model's performance across our different types of inputs.

In this subsection, we will outline certain relevant steps for improvement in the writing of prompts and pseudo-codes, the enhancements of each type of prompt and our expectations about how ChatGPT can deal with each one of them.

*Natural Language prompt*

> Create a meal plan for a paleolithic diet for each day of the week, considering a budget of 50 dollars per week, aiming for lean muscle gain, and including a detailed shopping list. Additionally, explain the nutritional benefits of each chosen meal.

In this prompt, we can visualize a multi-intentional interaction, where it is expected that ChatGPT will be capable of:

1. Planning the weekly meals, taking into account the indicated dietary habits and explaining the nutritional benefits of each meal.
2. Creating a shopping list, taking into consideration budgetary restrictions.

*Enhancement of Prompt in NL*

Once the intentions and prompt are defined, we will enhance it, exploring one of the main characteristics of Generative Artificial Intelligence (IAGen), the generation of personalized responses (Jovanovic & Campbell, 2022). Based on this principle, we make alterations so that, when inserting the prompt into ChatGPT, the user is questioned and provides information for the LLM to develop the objectives in a personalized and customized manner.

As we see in research that addresses the use of prompts in ChatGPT both in the field of education (Mollick & Mollick, 2023), in medicine (Heston & Khun, 2023), and in the financial field (Yue et al., 2023), the focus is on developing prompts that allow the user to insert specific information. Thus, exploring one of the specificities of chatbots, the interaction with the user (Heston & Khun, 2023).

Based on these principles, we model the interactive prompt (Wang et al., 2023) so that, after the input, ChatGPT can request information from the user and incorporate their data to ensure the completion of the requested process, similar to the work of Mollick & Mollick (2023). The incorporation of the elements described above led us to restructure the prompt aiming for generalization; understood by us as the characteristic that makes the prompt usable with different variables and the train of thought, in this way the prompt takes the form:

> Create a meal plan for each weekday based on the user's information, inquire about the diet the user follows, ask about the goal of the diet, inquire about the available budget, make a detailed shopping list, and explain the nutritional benefits of each chosen meal.

The process of prompt development, even in natural language, can be enhanced by following a train of thought, as highlighted by Wei et al. (2022). This approach, known as 'Chain of Thought', involves breaking down and organizing natural language into a series of sequential steps. By doing so, it guides the language model through a step-by-step reasoning process, mirroring how humans solve complex problems. Essentially, this technique not only facilitates the decomposition of multifaceted tasks into more manageable components but also increases the transparency and interpretability of the model's reasoning process. Moreover, the 'Chain of Thought' proves particularly effective in enhancing the ability of models to deal with problems requiring more detailed and nuanced reasoning, as demonstrated in empirical studies covering areas from arithmetic to symbolic reasoning. Applying such principles to our prompt, we realize that it takes the following form:

> To develop a meal plan, inquire about the diet the user follows, ask about the user's goal with the diet, inquire about the available budget, create a meal plan for each weekday based on the user's information, and explain the nutritional benefits of each chosen meal. Make a detailed shopping list with the necessary foods for preparing the meals and the user's budget.

We notice here a logical organization that first takes into account the user's diet and financial conditions before developing the meal plan. The request for the explanation of nutritional benefits was placed closer to the "meal plan" so that the nutritional information accompanies the plan and not the shopping list. Through the prompt, we reinforce the idea that the shopping list should bring the necessary foods for preparing the meals, intending that the LLM adheres to such items.

We will also adopt one of the techniques brought by prompt engineering (Heston & Khun, 2023), in which the prompt requests that the chatbots assume a certain role. Fulfilling such a request provides a context to the situation, which can guide the Generative AI to the desired pathway. Relating the technique and articulating the prompt, we have the following result.

> Act as a nutritionist who will develop a meal plan, inquire about the diet the user follows, ask about the goal the user has with the diet, inquire about the available budget, create a meal plan for each weekday based on the user's information, and explain the nutritional benefits of each chosen meal. Now act as a domestic economy specialist and make a detailed shopping list with the necessary foods for preparing the meals and the user's budget.

With the prompt structured, contextualized, and organized, we will move on to incorporating emotion into the text of the prompt (Li et al., 2023). Using the emotional intelligence present in the LLM, it is possible to enhance its performance by adding certain sentences. Thus, within the context of the prompt, which deals with the user's diet, we will include the phrase "The user's health depends on the results," leading the prompt to take the following form:

> Act as a nutritionist who will develop a meal plan, inquire about the diet the user follows, ask about the goal the user has with the diet, inquire about the available budget, create a meal plan for each weekday based on the user's information, and explain the nutritional benefits of each chosen meal. Now act as a domestic economy specialist and make a detailed shopping list with the necessary foods for preparing the meals and the user's budget. The user's health and budget depend on the execution of these tasks.

With these enhancements, we finalize the development of the prompt in natural language (LN). This prompt will be used to generate information for the 'Natural Language' analysis unit. We will subsequently present the parameters to be inserted in the interactions with the chatbots, such as the type of diet and available budget.

*Pseudocoding*

In the context of this article, pseudocoding represents a crucial step in analyzing the interaction with ChatGPT. Pseudo-code, as described by Calzadilla (2018), is a powerful tool for structuring complex solutions in a clear and comprehensible manner, even for those who do not have extensive programming experience. It uses language close to natural, with easily recognizable terms and actions, and a symbolism inspired by algebra and mathematical logic, making the underlying logic to the instructions more accessible.

In this subsection, we will transform the prompt developed in the previous chapter into pseudo-code similar to Python coding. This transformation aims to demonstrate how the structure and clarity provided by pseudo-code can influence the accuracy and efficiency of ChatGPT's responses. Pseudocoding will serve as a means to organize and detail the steps and conditions required for ChatGPT to execute the tasks designated in the prompt.

Even if there may not be a need for mastery of Python language, it is necessary to be familiar with some of the programming structures of this software. The first of these are the "Keywords", which indicate a specific action (Rossum, 2023), and the structures of lines. It is interesting to differentiate functions from comments, as comments will only serve to guide the work of the LLM, but not exactly defining its function.

First, we will separate each of the functions present in the text into individual lines. Then, we will number them so that the LLM can follow them strictly. We will connect them with the commands "ACT", "INQUIRE", "CREATE". Incorporating these concepts into the prompt, its structure in pseudo-code takes the following format:

> 1) ACT as a nutritionist who will develop a meal plan;
>
> 2) INQUIRE about the diet the user follows;
>
> 3) INQUIRE about the intended goal with the diet;
>
> 4) INQUIRE about the available budget;
>
> 5) Using the responses from 2, 3, and 4, CREATE a meal plan for each weekday that explains the nutritional benefits of each chosen meal;
>
> 6) ACT as a domestic economy specialist;
>
> 7) Using the response from 4, CREATE a shopping list with the necessary foods for preparing the meals.
>
> 8) The user's health depends on this task.

The prompt presented above will be the prompt entered into ChatGPT for generating responses for the 'Pseudo-code' analysis unit. Along with the prompt, the usage parameters will be inserted, which are presented in the following session.

*What we expect of this different type of inputs?*

In the exploration of ChatGPT's interactions with varied input forms, our anticipation pivots on understanding how different types of prompts shapes the model's performance. Drawing from our discussions, this subsection delves into the expected outcomes, potential advantages, or disadvantages tied to each input type, alongside contemplating the technical feasibility and inherent challenges of enhancing interactions with ChatGPT.

Natural language inputs, characterized by their intuitive and user-friendly nature, are presumed to facilitate broader accessibility, allowing users with minimal technical background to interact with ChatGPT effectively (Wolfram, 2023). The hypothesis suggests that while natural language prompts offer ease of use, they might lead to ambiguities that challenge the model's precision in understanding and executing user intentions. The inherent vagueness and the multiplicity of interpretations available in natural language could potentially hinder the model's performance in generating targeted responses, highlighting a trade-off between accessibility and precision.

Pseudo-code input, with their inherent structure and logic, stand at the forefront of enhancing ChatGPT's precision in query comprehension and response generation. These inputs, by embodying the essence of programming logic, are designed to significantly reduce ambiguity, steering ChatGPT toward outputs that are not only precise but also aligned with the functional expectations of the query (Kusal et al., 2022; Wu & Feng, 2018). This precision, rooted in the structured nature of pseudo-code, facilitates a clearer communication channel between the user and ChatGPT, ensuring that the model's responses are directly relevant and applicable to the task at hand.

Furthermore, the endeavor to refine ChatGPT's processing of natural language inputs, aiming to mitigate ambiguities while nurturing the model's inventive output, complements the advancements made with pseudo-code. It entails a sophisticated recalibration of the model's interpretive algorithms, striking a delicate balance that respects user intentions while fostering ChatGPT's inherent capacity for generating creative and contextually rich responses (Wei et al., 2022).

In essence, the exploration of the pseudo-code inputs can reveal a compelling landscape of advantages that these structured inputs. It can not only promise to elevate ChatGPT's operational accuracy but also open avenues for a more defined, logical, and goal-oriented interaction framework. While maintaining a vigilant stance on the challenges and technical intricacies involved, the strategic emphasis on pseudo-code inputs heralds a promising direction for enriching ChatGPT's interface with human language.

### Definition of Parameters for Data Generation

For this study, we will use the Chat GPT 4.0 turbo version, which includes the DALL.E, web browsing, and analysis functions in its guidelines. To ensure parallelism between the units of analysis, we defined standardized responses to answer the questions that the respective chatbots will ask the user. The choice of the Paleolithic diet, lean muscle gain as a fitness goal, and a budget of 50 dollars per week were deliberate and strategic, designed to mirror realistic scenarios that users might input into ChatGPT. This selection was guided by several key considerations:

**Diet:** The Paleolithic diet was selected due to its popularity and specificity. It represents a modern nutritional plan based on the presumed diet of early humans, focusing on whole foods, and excluding processed foods, grains, and dairy (Frączek et al., 2021). This diet's restrictive nature makes it a pertinent choice for testing ChatGPT's capability to generate meal plans that adhere to specific dietary frameworks. It challenges the model's ability to understand and apply nutritional principles within the constraints of a predefined dietary pattern, aligning with the study's goal to evaluate the adaptability and precision of ChatGPT in generating customized advice.

**Goal:** The objective of lean muscle gain complements the chosen diet by introducing a common fitness goal that requires precise nutritional strategies (Altyar, 2020). This goal necessitates a balance of macronutrients to support muscle development while adhering to the dietary restrictions of the Paleolithic framework. It tests ChatGPT's capacity to tailor nutritional recommendations that not only fit dietary constraints but also support specific fitness objectives, showcasing the model's potential in providing nuanced and goal-oriented guidance.

**Budget:** Establishing a budget constraint introduces a layer of complexity to the task, simulating a common real-world consideration for individuals following specific diet plans. It requires ChatGPT to optimize meal planning within financial limitations, reflecting the model's ability to incorporate economic factors into its recommendations. This parameter is crucial for evaluating ChatGPT's effectiveness in generating practical and economically feasible meal plans, ensuring that the proposed solutions are not only nutritionally adequate but also accessible to users with budgetary considerations.

By setting these parameters, the study aims to explore ChatGPT's proficiency in navigating the intricacies of diet planning, nutritional optimization, and budget management. These specific choices align with the study's objectives to assess the depth, accuracy, and practicality of ChatGPT's responses to complex, multi-dimensional queries, providing a comprehensive evaluation of its capabilities in delivering personalized and contextually relevant advice.

To view the dialogs generated using the selected LLM for the study, see appendices C, D and E.

## DATA ANALYSIS

In this subsection, we will conduct a careful analysis of the interactions with ChatGPT, applying the discourse analysis method (Jørgensen & Phillips, 2011). Our goal is to understand ChatGPT's responses to different types of inputs, such as natural language and pseudo-codes. We will focus on identifying patterns, contradictions, and unique elements in the responses, assessing the efficiency and accuracy of ChatGPT in various contexts. This analysis seeks to reveal the limits and capabilities of the model, providing significant insights into human-computer interaction in the field of conversational artificial intelligence.

In Figure 1, we present the results from the analysis of the units in each of the four selected categories. In the graphics it's possible to visualize the unit attempted in every criterion of the categories. More details, such as elements that led to the inference of each classification.

**Figure 1**

*Category 1: Understanding of Intentions*

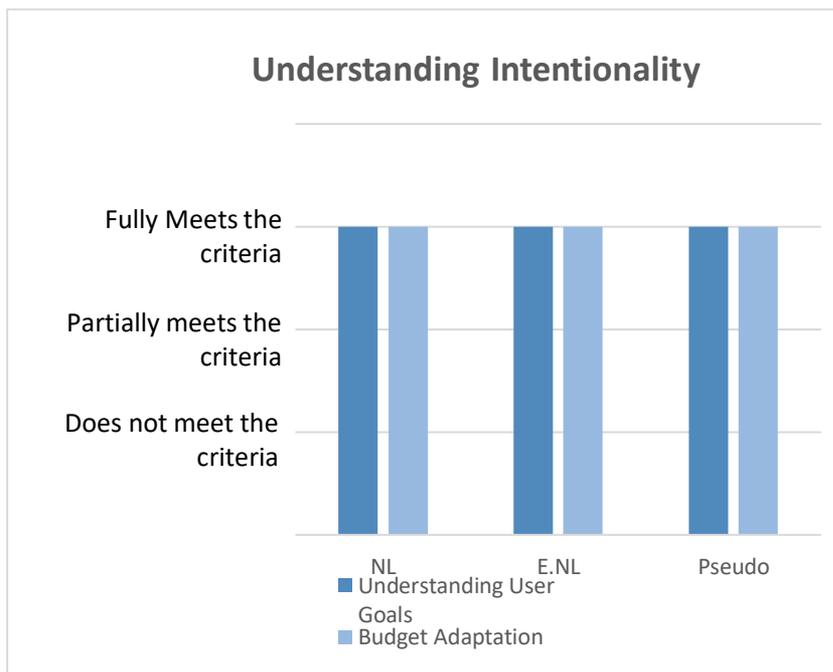

In the analysis of Understanding Intentionality across various units of ChatGPT, the results demonstrate a consistent proficiency in identifying and adhering to user intentions. Unit A (Natural Language) and Unit B (Enhanced Natural Language) both accurately recognized and addressed the user's objective of developing a Paleo diet-based meal plan focused on lean mass gain, within a specified budget. These units not only grasped the main goal but also incorporated detailed pricing information, directly aligning their responses with the user's financial constraints.

Unit C (Pseudo-code) effectively met the core request for a budget-friendly meal plan. However, unlike Units A and B, which provided detailed pricing, Unit C offered general cost-saving tips without specific price details. This distinction suggests a difference in approach to integrating budget considerations into the meal planning process.

This examination indicates ChatGPT's capability to comprehend and execute requests across different forms of input, from natural to structured languages, with a particular emphasis on understanding user goals and adapting responses to budget constraints. The model consistently fulfills the primary requirements of the tasks, demonstrating its adaptability and accuracy in responding to varied user intents.

In examining interpretability across different units (Figure 2), there's a notable differentiation in how each unit structures and presents its responses to facilitate user comprehension. Unit A, employing natural language, partially aligns with the criterion, integrating nutritional benefits close to meal descriptions, which, while informative, may slightly disrupt the flow of the meal plan presentation. Conversely, Units B and C, employing enhanced natural language and pseudo-code respectively, fully adhere to the criteria, showcasing a clear and logical arrangement of nutritional information alongside meal plans, thereby enhancing user understanding.

**Figure 2**

*Category 2: Interpretability*

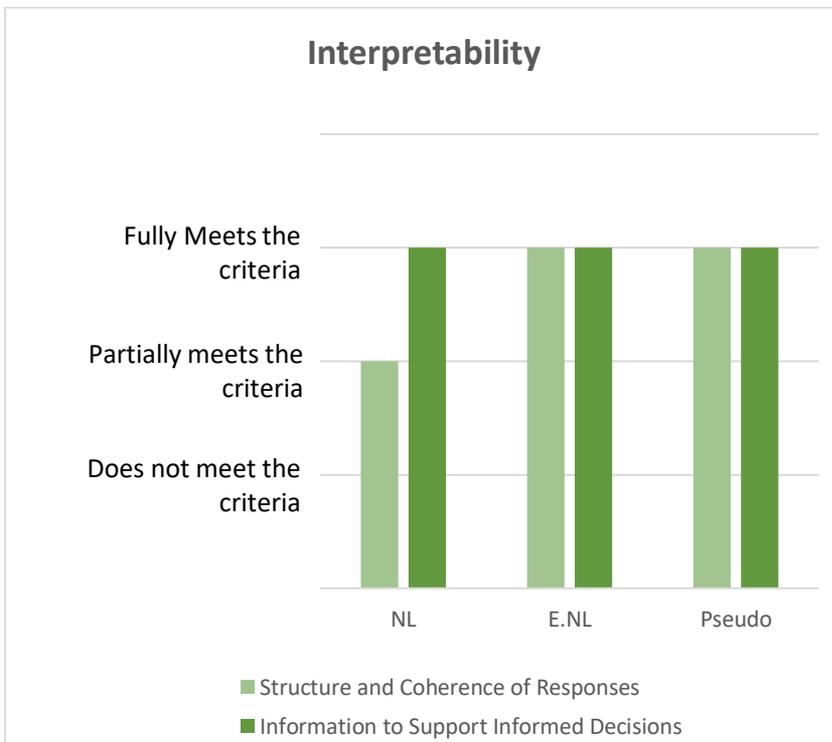

In the category of Completeness, which evaluates the thoroughness of ChatGPT's meal plans, our analysis scrutinized how each unit tackled the integration of daily dietary requirements (Figure 3). Comprehensive Coverage of Needs was the first criterion examined, assessing the all-encompassing nature of the meal plans provided by each unit. Natural Language (Unit A) partially met this criterion by covering all main meals and snacks, yet it lacked in offering a broad variety within each meal category. Conversely, Enhanced Natural Language (Unit B) and Pseudo-code (Unit C) fully satisfied this criterion by presenting complete and balanced dietary plans, with Unit B standing out for its diverse range of options.

**Figure 3**

*Category 3: Completeness:*

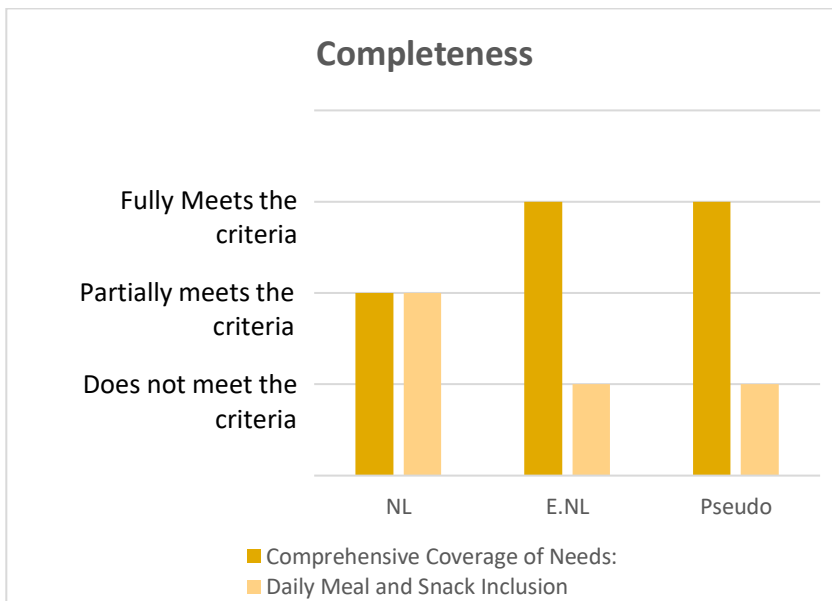

For the criterion of Daily Meal and Snack Inclusion, which looked at the inclusion of diverse meal types throughout a day, Natural Language (Unit A) only partially met this, offering limited snack options. Enhanced Natural Language (Unit B) and Pseudo-code (Unit C) did not meet the criteria, as neither provided snack suggestions in their daily plans.

In the Creativity category, the focus was on the language model's innovation in generating meal plans (Figure 4). Meals Development Creativity was our first lens, evaluating the diversity and quantity of meal options proposed. Natural Language (Unit A) did not meet this criterion due to its limited variety, offering only four meals for the entire week. In contrast, Enhanced Natural Language (Unit B) and Pseudo-code (Unit C) excelled, each developing fifteen diverse meals, thereby fully meeting the criterion.

For Adaptability and Flexibility of Meal Plan, the assessment looked at the model's ability to suggest ingredient substitutions or meal variations. Natural Language (Unit A) failed to meet this criterion, providing a rigid set of meals with no evident flexibility. Enhanced Natural Language (Unit B) and Pseudo-code (Unit C) partially met the criteria by indicating the possibility of adjusting meals based on personal preference, though they presented fixed meal plans.

**Figure 4**

*Category 4: Creativity*

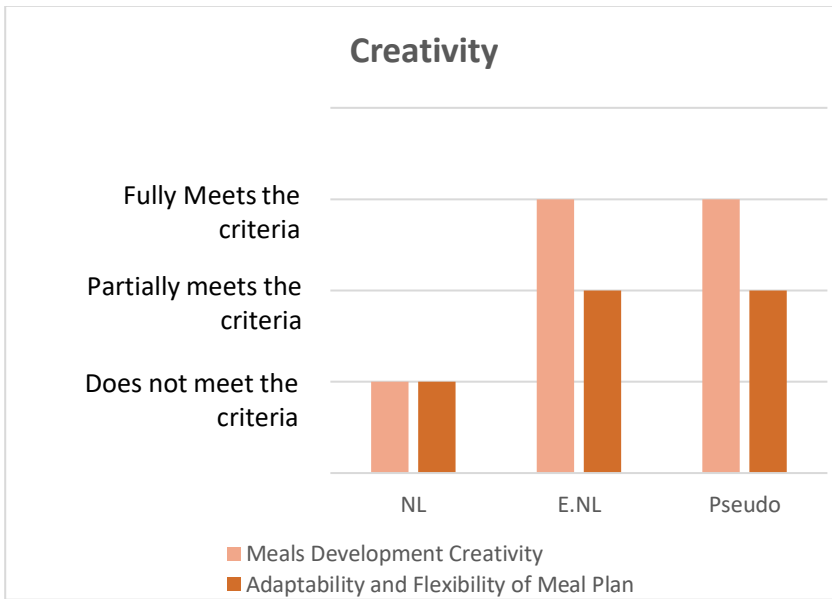

**Discussion of Results**

    The definition of the four categories enabled a comprehensive understanding of the model's interaction with the form of language input by the user. We will analyze the nuances brought by each category together at first graphically, visible in the Figure 5 of integrated results.

**Figure 5**

*Overall results*

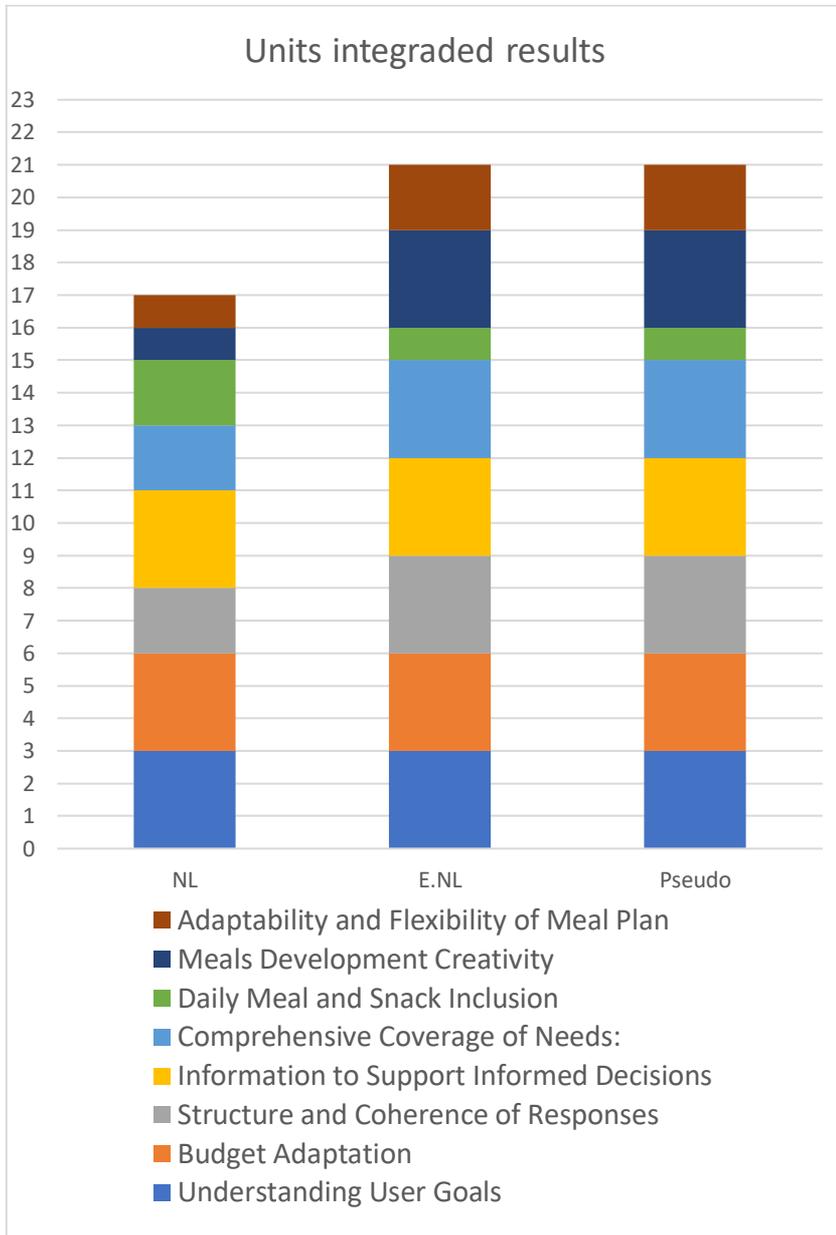

To enhance the clarity and academic rigor of the analysis, we leveraged numerical evaluation methods, as detailed in a preceding chapter. The graphical representation succinctly illustrates the comparative performance of the different input methods. It reveals that while Natural Language displayed a relatively lower performance, both Enhanced Natural Language and Pseudo-code exhibited superior capabilities.

In examining the "Understanding Intentionality" category, it is evident from the Figure 1 that all three units attained an equivalent performance level across the defined criteria. This uniformity suggests that ChatGPT's ability to comprehend and accurately fulfill user requests is not contingent on the input format. Through this category and its specific criteria, we observe that ChatGPT's response effectiveness is consistent across different user inputs. This insight shifts the focus to evaluating how ChatGPT processes these requests and responds to varied types of inputs.

When contrasting natural language with its enhanced counterpart, which incorporates prompt engineering techniques (Heston & Khun, 2023) and the chain of thoughts methodology (Ling et al., 2023; Wei et al., 2022), as well as emotional interaction with the chat (Li et al., 2023), notable advantages are observed. In terms of Interpretability and Creativity, Enhanced Natural Language distinguishes itself by presenting a meal plan with a wider variety of options and a more coherent informational flow. Our hypothesis suggested that while Natural Language is more accessible, the potential absence of guidance or the misplacement of requests within the structure of this input could contribute to significant disparities in Creativity and Interpretability. Through the application of prompt engineering (Heston & Khun, 2023), specific instructions for the tasks were inferred, thereby providing more targeted guidance for the LLM's vectorization process, resulting in more inventive responses. Furthermore, the adoption of the chain of thought approach (Ling et al., 2023a; Wei et al., 2022) enhanced the organization, as evidenced in Criterion 2.A, Structure and Coherence of Responses. The responses were methodically segmented by the LLM, organizing the answers in alignment with the sequence of instructions.

In engaging with ChatGPT through a code-resembling language, specifically pseudo-code, referred to as Unit C, we incorporated chain of thought techniques (Ling et al., 2023a, 2023b; Wei et al., 2022) and prompt engineering (Heston & Khun, 2023) to formulate our pseudo-code. This approach yielded concise text that was expedient to construct. Despite its brevity, this unit exhibited commendable performance in our evaluations, paralleling that of the Enhanced Natural Language. Regarding Interpretability, both units excelled, fully meeting the criteria with their structured and coherent responses. They adeptly presented detailed nutritional information in a logically organized manner, thereby enhancing user comprehension, and facilitating informed decision-making. This achievement reflects the model's capacity to process and articulate information consistent with user expectations, reinforcing the significance of clarity and user understanding in human-computer interaction, as highlighted in the literature (Kusal et al., 2022; Wu & Feng, 2018). Concerning Completeness, both units were highly effective, encompassing all necessary elements of a meal plan. This observation corroborates Lee & Choi's (2003) emphasis on the critical role of comprehensive knowledge creation and dissemination within organizational contexts, principles equally applicable to the model's generation of complete and thorough responses. In assessing creativity, both units demonstrated equivalent levels of performance across criteria, showcasing their innovative capabilities in meal planning.

Leveraging the pseudo-code approach enabled us to achieve the requisite simplicity. By employing an input method that can severally decrease the potential for nuances in language and clearly and systematically delineates the sequence of functions or requirements, we devised a more efficient means to elicit responses comparable to those obtained from the Enhanced Natural Language using our Pseudo-code. Regarding simplicity, Pseudo-code demonstrated an easier and quicker method for composing very efficient prompts.

## CONCLUSION

Throughout this study, our objective was to explore and elucidate the distinctions among responses generated by ChatGPT when interfacing with various forms of inputs. The initial goals included examining ChatGPT's capability to process multiple intentions and input formats, extending from natural language to more structured forms, such as pseudo-code. The methodology employed, integrating case study approaches (Yin, 2001) with discourse analysis (Jørgensen & Phillips, 2011), facilitated a thorough and contextual investigation of these interactions.

The findings of this study elucidate how distinct forms of inputs uniquely influence the ChatGPT performance. Natural language, despite its accessibility and intuitiveness, often yielded responses that were limited in interpretability and lacked creative innovation. Conversely, enhanced natural language, augmented through sophisticated prompt engineering and chain of thought techniques, exhibited marked enhancements in both interpretability and creativity, rendering it a superior tool for tasks necessitating nuanced and inventive responses. The adoption of pseudo-code marked a shift towards a more structured mode of communication akin to programming languages, achieving parity in performance with Enhanced Natural Language.

The selection of the optimal language for interacting with ChatGPT ought to be contingent upon the user's specific context and requirements. It is critical to recognize that employing natural language, particularly with multiple intentions, may result in unpredictable and varied responses. In contrast, pseudo-code offers a structured approach that facilitates more explicit recognition of intentions by the language model, leading to clearer and more deterministic outcomes. This attribute is particularly advantageous in situations where precision and clarity in instructions are essential. Furthermore, the adoption of pseudo-code can enhance the determinism of interactions with ChatGPT, proving invaluable in contexts demanding consistent and predictable responses. Additionally, from a time efficiency perspective, pseudo-code can be composed more swiftly due to its reduced character count, offering a practical advantage in rapid response generation.

This article has delineated the operational mechanics of ChatGPT, tracing the journey from user input to the language model's generated output. We have meticulously explored various forms of language input and examined the nuanced responses elicited by each. It is evident that every form of language possesses its unique characteristics and should be employed in alignment with the user's specific intentions.

In this research, we have advanced the methodology of employing pseudo-codes, observing their capacity to generate results that are more deterministic, concise, and definitive. We introduced the concept of Pseudo-code Engineering, where a natural language prompt undergoes transformation into pseudo-code. The comprehensive analysis underscores that Pseudo-code emerge as optimal choices for formulating prompts for ChatGPT.

In conclusion, the exploration of Pseudo-code Engineering within our study illuminates a promising avenue for refining interactions with Large Language Models like ChatGPT, offering a spectrum of future applications spanning from educational tools to sophisticated software development. Our results encourage subsequent research to assess the predictability and stability of AI responses when interfaced through pseudo-code, emphasizing the need for a structured approach to enhance AI comprehension and output. The concept of Pseudo-code Engineering not only aims to streamline AI communications but also seeks to unlock a new realm of possibilities in human-AI interaction, making it more precise, efficient, and accessible across varied sectors. Anticipating further experimentation and analysis, we remain committed to investigating the full potential of pseudo-codes in shaping the future of AI technologies, thereby ensuring their expanded utility and effectiveness.

This investigation illuminates ChatGPT's proficiencies and constraints in processing diverse linguistic inputs, providing indispensable insights for users aiming to optimize their engagement with this sophisticated language model. The adoption of pseudo-codes holds significance across various domains, from educational endeavors to software development, heralding new avenues for research and practical implementation.

## ACKNOWLEDGEMENTS

This study was financed in part by the Coordenação de Aperfeiçoamento de Pessoal de Nível Superior - Brasil (CAPES) - Finance Code 001.

# APENDICES

## Unit A – Natural Language

Available at https://chat.openai.com/share/d99607f7-b2af-4fe5-9208-5211b54c401d

### Unit B – Enhanced Natural Language

Available at: https://chat.openai.com/share/f92051f4-0ffe-4f53-ba3a-e31f3805cd2d

**ChatGPT 4**

Tuesday
- **Breakfast**: Smoothie with banana, almond milk, and a scoop of Paleo-friendly protein powder.
- **Lunch**: Beef stir-fry with bell peppers, broccoli, and carrots.
- **Dinner**: Roasted turkey breast with sweet potato and green beans.

Wednesday
- **Breakfast**: Omelette with bell peppers, onions, and a side of fresh fruit.
- **Lunch**: Tuna salad with mixed greens, cucumber, and olives.
- **Dinner**: Pork chops with roasted Brussels sprouts and apple slices.

Thursday
- **Breakfast**: Paleo pancakes made with almond flour, topped with fresh berries.
- **Lunch**: Chicken lettuce wraps with avocado and tomato.
- **Dinner**: Grilled shrimp with a mixed vegetable salad.

Friday
- **Breakfast**: Fruit salad with a variety of seasonal fruits.

Message ChatGPT...

ChatGPT can make mistakes. Consider checking important information.

**ChatGPT 4**

Friday
- **Breakfast**: Fruit salad with a variety of seasonal fruits.
- **Lunch**: Beef and vegetable skewers.
- **Dinner**: Baked cod with a side of roasted cauliflower and a mixed green salad.

### Nutritional Benefits
- **Lean Proteins** (chicken, turkey, salmon, tuna, shrimp, beef): Essential for muscle growth and repair.
- **Vegetables and Fruits**: Provide vitamins, minerals, and fiber for overall health and digestion.
- **Healthy Fats** (avocado, nuts): Support hormone production, crucial for muscle growth.

### Shopping List (Within $50 Budget)
- **Proteins**: Chicken breast, salmon, turkey breast, canned tuna, pork chops, shrimp, ground beef, cod.
- **Vegetables**: Spinach, mixed greens, cherry tomatoes, asparagus, bell peppers, broccoli, carrots, sweet potato, green beans, Brussels sprouts, lettuce, cucumber,

Message ChatGPT...

ChatGPT can make mistakes. Consider checking important information.

[screenshot of ChatGPT 4 conversation showing meal plan details: Proteins, Vegetables, Fruits, Others lists, Tips for Budget, and closing remarks]

### Unit C – Enhanced Natural Language

Available at: https://chat.openai.com/share/1923921c-5042-4c5e-9015-061c7332e61c

[screenshot of ChatGPT 4 conversation with prompt:
1) ACT as a nutritionist who will develop a meal plan;
2) QUESTION about the diet the user follows;
3) QUESTION about the intended objective of the diet;
4) ASK about the available budget;
5) Using answers 2, 3 and 4 CREATE a meal plan for each working day that explains the nutritional benefits of each meal chosen;
6) ACT as a home economics expert;
7) Using the answer from 4 CREATE a shopping list with the foods needed to prepare meals.
8) The user's health depends on this task

ChatGPT response: Absolutely, I can assist you with creating a meal plan and shopping list. Let's start by gathering some necessary information:
1. **Dietary Preferences and Restrictions**: Could you please tell me about the diet you currently follow? For example, are you vegetarian, vegan, gluten-free, or do you have any food allergies or preferences?]